%% file: main_arxiv.tex
\PassOptionsToPackage{table}{xcolor}
\documentclass[11pt]{article}

\usepackage[preprint]{acl}
\usepackage{lineno}
\usepackage{times}
\usepackage{latexsym}
\usepackage[T1]{fontenc}
\usepackage[utf8]{inputenc}
\usepackage{microtype}
\usepackage{inconsolata}
\usepackage{graphicx}
\usepackage{booktabs}
\usepackage{tabularx}
\usepackage{array}
\usepackage{xcolor}
\usepackage[skins]{tcolorbox}
\usepackage{fontawesome5}
\usepackage{enumitem}
\usepackage{dblfloatfix}
\usepackage{cuted}
\usepackage{longtable}
\usepackage{supertabular}
\usepackage{needspace}
\usepackage{placeins}
\usepackage{tikz}
\usetikzlibrary{arrows.meta,calc,positioning}

\definecolor{specshade}{HTML}{F5E8D0}
\definecolor{repshade}{HTML}{DDEBF7}
\definecolor{orchshade}{HTML}{E9E3F4}
\definecolor{regshade}{HTML}{E3F1E6}
\definecolor{couplingshade}{HTML}{F6DEDE}
\definecolor{panoramarow}{HTML}{F5F4F1}
\definecolor{specchip}{HTML}{E8C98F}
\definecolor{repchip}{HTML}{AFCFE7}
\definecolor{orchchip}{HTML}{CCBEE3}
\definecolor{regchip}{HTML}{B9DCBF}
\definecolor{couplingchip}{HTML}{EAB6B6}
\definecolor{TakeawayLine}{RGB}{21,67,118}
\definecolor{TakeawayBg}{RGB}{248,250,252}

\newtcolorbox{focusbox}[1][]{
  enhanced,
  frame hidden,
  boxrule=0pt,
  sharp corners,
  borderline west={2.5pt}{0pt}{TakeawayLine},
  colback=TakeawayBg,
  top=1.5mm,
  bottom=1.5mm,
  left=2.5mm,
  right=1.5mm,
  before skip=1.5mm,
  after skip=1.5mm,
  fontupper=\small\linespread{0.95}\selectfont,
  before upper={\textbf{\sffamily #1}\\[0.8mm]}
}

\newtcolorbox{keydistinctionbox}[1][]{
  enhanced,
  colback=TakeawayLine!4,
  colframe=TakeawayLine!45,
  boxrule=0.45pt,
  sharp corners,
  borderline west={3pt}{0pt}{TakeawayLine},
  top=1.8mm,
  bottom=1.8mm,
  left=2.7mm,
  right=1.8mm,
  before skip=2mm,
  after skip=2mm,
  fontupper=\small\linespread{0.96}\selectfont,
  before upper={\textcolor{TakeawayLine}{\textbf{\sffamily #1}}\\[0.9mm]}
}

\newcolumntype{Y}{>{\raggedright\arraybackslash}X}
\newcolumntype{P}[1]{>{\raggedright\arraybackslash}p{#1}}
\newcommand{\role}[1]{\textsc{#1}}

\title{What Language Does and What the Evidence Supports:\\
A Functional Role Taxonomy and Evidence Audit of Language Grounding in Embodied Agents}
\author{
  \textbf{Yifan Guo\textsuperscript{1}},
  \textbf{Chenghao Li\textsuperscript{2}},
  \textbf{Zhu Wang\textsuperscript{1,\textdagger}},
  \textbf{Wei Xu\textsuperscript{1}},
  \textbf{Yu Li\textsuperscript{1}},
  \textbf{Yulong Zhu\textsuperscript{1}},
  \textbf{Zhuo Sun\textsuperscript{1}},
\\
  \textbf{Bin Guo\textsuperscript{1}},
  \textbf{Zhiwen Yu\textsuperscript{1}}
\\
\\
  \textsuperscript{1}Northwestern Polytechnical University \\
  \textsuperscript{2}University of Electronic Science and Technology of China \\
  \textsuperscript{\textdagger}Corresponding author.
}

\begin{document}
\maketitle

\input{sections/00_abstract}
\input{sections/01_introduction}
\input{sections/02_scope_related}
\input{sections/03_framework}
\input{sections/04_functional_roles}
\input{sections/05_evidence_audit}
\input{sections/06_implications}
\input{sections/07_conclusion}
\input{sections/08_limitations}
\input{sections/09_ethical_considerations}

\bibliography{references}

\appendix
\input{appendices/appendices}

\end{document}

%% file: sections/00_abstract.tex
\begin{abstract}
Foundation models place language throughout embodied agents, but its presence does
not show what it contributes or how well that contribution is grounded. This survey
separates these two questions. We define five non-exclusive functional roles for
language: Specification, Embodied Representation, Action Orchestration, Grounding
Regulation, and Execution Coupling. For each role, we trace the path from linguistic
content to its embodied consumer and identify the observations or interventions that
can test the claimed responsibility. Applying this framework to the reviewed
literature reveals a recurring gap between functional use and evidential support.
Interpretable or revised linguistic intermediates may be incorrect, go unused, or fail
to affect later behavior. Even when actions are directly conditioned on language,
system-level success does not by itself isolate language's contribution. We therefore
evaluate grounding claim by claim, asking whether the reported evidence supports the
specific responsibility assigned to language. Using role claims rather than
architectures as the unit of comparison allows us to compare modular and end-to-end
embodied agents without extending conclusions beyond the reported evidence.
\end{abstract}

%% file: sections/01_introduction.tex
\section{Introduction}
\label{sec:introduction}

Language now participates throughout embodied behavior. It can define a task or
organize learned skills \citep{ahn2022icanisay,liang2023code}. Language-aligned
representations can make spatial state available to later decisions
\citep{huang2023visual}. Language can also express a reward or revise behavior after a
detected failure
\citep{yu2023language,duan2024ahavisionlanguagemodeldetectingreasoning}. In other
systems, it directly conditions the action policy \citep{zitkovich2023rt}. These
mechanisms are commonly discussed under the heading of \emph{language grounding},
although they assign different responsibilities to linguistic content.
Improved language-conditioned success, however, does not establish that linguistic
content is better grounded. Better perception or broader robot data may explain the
gain. Changes to the action model, planner, or controller can do so as well
\citep{zitkovich2023rt,li2024cogactfoundationalvisionlanguageactionmodel}. The evidence
must therefore distinguish the linguistic contribution from these alternatives.

\input{figures/figure_intro_audit}

Grounding can be evaluated only after the responsibility assigned to language has been
identified. Our central question is therefore: \textbf{\emph{What functional roles does
language play in embodied agents, and what embodied evidence warrants each role?}} We
distinguish five non-exclusive roles: Specification, Embodied Representation, Action
Orchestration, Grounding Regulation, and Execution Coupling. These roles may coexist in one system. We therefore characterize a paper by the roles
instantiated by its language component rather than force the entire system into a single
category. For each role, we trace the path from linguistic content to the embodied
component that consumes it. We then ask what reported observation or intervention can
test the responsibility assigned to language. Figure~\ref{fig:intro-audit} illustrates
this form of analysis.

Applying this role-based view reveals a recurring problem: evidence for one property of
a system is sometimes used to support a stronger claim. We call this an
\emph{evidential substitution}. Four forms recur in the reviewed literature.
\begin{enumerate}[leftmargin=1.5em,itemsep=1pt,topsep=3pt,parsep=0pt]
    \item \textbf{Explicit language is not evidence of grounded content.} A readable
    intermediate makes a language role easier to inspect. It does not show that the
    content matches the current environment or that the agent relies on it when acting.
    \item \textbf{Internal revision is not embodied correction.} A critique or revised
    plan shows that an internal state has changed. It counts as embodied correction only
    if it changes a later executable decision. A recovery claim additionally requires
    evidence of an improved outcome.
    \item \textbf{Task success is not language attribution.} Task success evaluates the
    complete system. Attributing a gain to language requires a comparison that directly
    targets the claimed language role.
    \item \textbf{Closeness to action is not stronger grounding.} Direct conditioning of
    action places linguistic content near motor output, but it does not isolate the
    linguistic contribution. Grounding strength depends on the evidence that bears on
    the claimed role.
\end{enumerate}

The audit does not assume that language contributions cannot be isolated. Many papers
include matched controls that test a particular role-bearing quantity. Far fewer trace
an observed outcome through revision to a changed executable attempt. In our analysis,
functional responsibility and evidential warrant are recorded separately. The first
identifies what language changes; the second determines what the reported experiments
establish about that change.

This review makes three contributions. \textbf{First}, it introduces a functional taxonomy that
represents embodied systems through non-exclusive language-role profiles rather than
architecture or processing stage. \textbf{Second}, it compares grounding mechanisms by tracing
each claimed role to the embodied tests that bear on it. \textbf{Third}, it audits the evidence
reported in the reviewed literature and turns recurrent gaps into concrete reporting
questions.

Section~\ref{sec:scope} defines the review scope.
Sections~\ref{sec:framework}--\ref{sec:audit} introduce the taxonomy, examine how the
five roles are implemented, and report the evidence audit.
Section~\ref{sec:implications} presents the implications for evaluation and reporting.
The appendices document corpus construction and provide the complete role and evidence
profiles.

%% file: figures/figure_intro_audit.tex
\begin{figure}[t]
\centering
\includegraphics[width=\columnwidth]{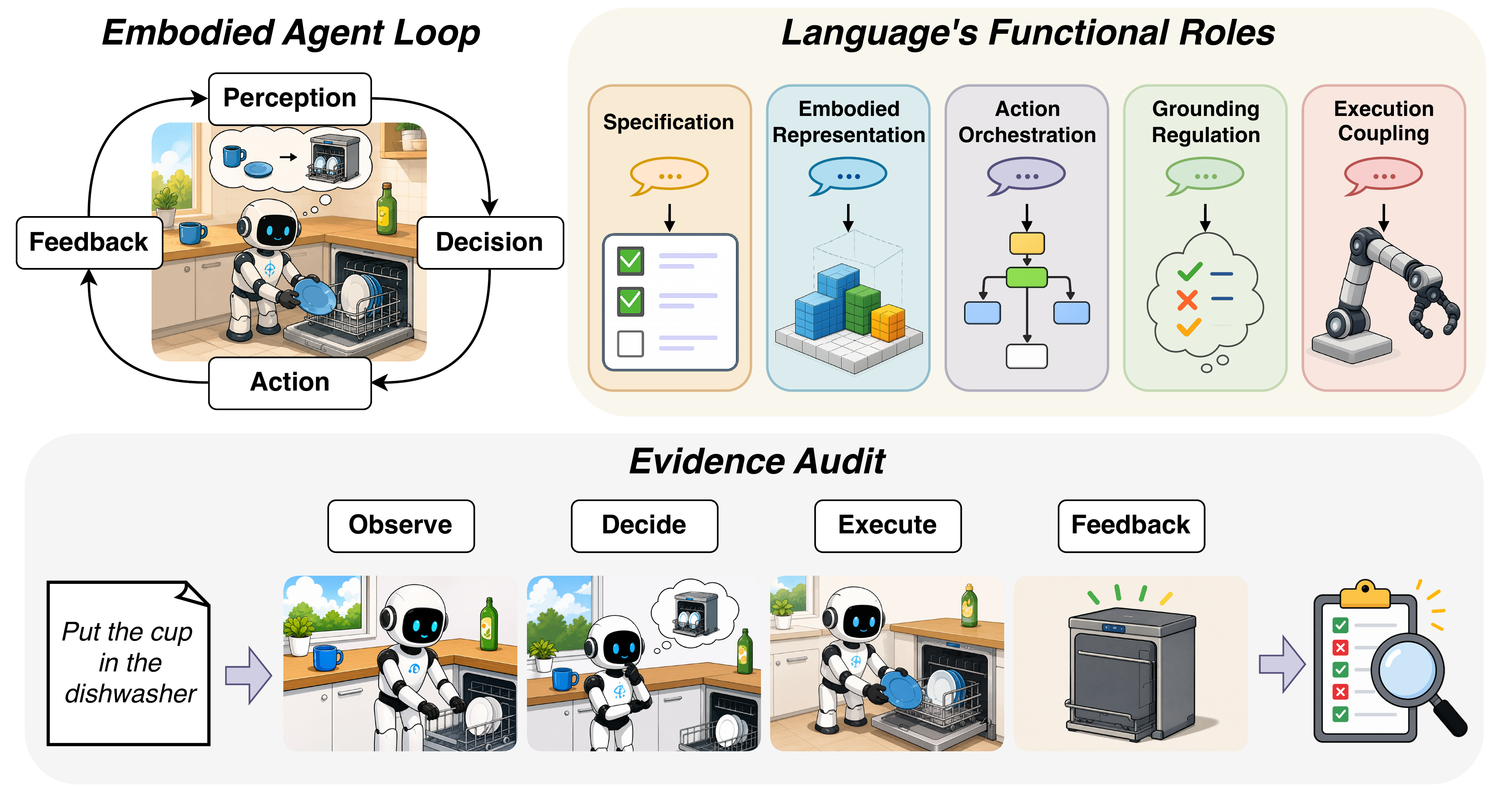}
\caption{Overview of the review framework. The embodied loop provides the context in
which language may assume five non-exclusive functional roles. Roles are assigned
according to what language-derived content does. For each paper--role claim, the
evidence audit asks whether the claimed contribution can be connected to downstream
embodied behavior.}
\label{fig:intro-audit}
\end{figure}

%% file: sections/02_scope_related.tex
\section{Scope and Review Positioning}
\label{sec:scope}

We define the review scope by what linguistic content changes in an embodied system,
not by the model label. The same rule applies to large language models (LLMs),
vision--language models (VLMs), multimodal large language models (MLLMs), and
vision--language--action models (VLAs). A paper is eligible only when linguistic
content has a reported path to task-relevant state or behavior in a situated agent.
This path may operate during the current interaction or through experience reused
later. Language used only for annotation, post hoc description, or offline scoring
does not qualify. We also exclude offline benchmarks with no agent--environment
consequence. These criteria determine inclusion, not evidential strength.
Continual-learning systems are included only when previously acquired
language-conditioned information has a reported effect on a later embodied loop;
continual learning is not treated as a separate functional role.
Appendix~\ref{app:review-protocol} documents the search and screening procedure.

This boundary retains the longstanding requirement that linguistic content remain
answerable to situated perception and action
\citep{harnad1990symbol,roy2005semiotic,bisk2020experience}.
Foundation-model-enabled agents make attribution difficult because language operates
inside a larger embodied system. A reported gain may come from components that
interpret the scene or produce motion. It may also depend on the data used to train
those components. Our question is therefore not whether the complete system succeeds,
but which change can be assigned to language and what evidence supports that
assignment.

Existing surveys commonly organize the field around model design, robot capabilities,
or learning paradigms
\citep{wang2024largelanguagemodelsrobotics,ma2026survey,chen2026semanticlifecycleembodiedai}.
These perspectives explain how embodied systems are built and what tasks they address.
We ask a different question: what responsibility does linguistic content assume, and
how is that responsibility tested? Our unit of comparison is the role-specific path
from linguistic content to behavior, which can be identified in both modular and
end-to-end systems. Because one paper may contain several such paths, the roles remain
non-exclusive. Appendix~\ref{app:survey-comparison} compares this analytical choice
with representative surveys.

%% file: sections/03_framework.tex
\section{From Language Function to Embodied Evidence}
\label{sec:framework}

Having defined the scope, we ask what difference linguistic content makes inside the embodied loop. The location of a language module does not answer this question; its downstream responsibility does. We distinguish five functional roles, summarized in Table~\ref{tab:roles}.

The roles are assigned to linguistic content rather than to the system as a whole or to a chronological stage. A planner may define the task and also select the skills used to pursue it. A spatial representation may describe embodied state while directly constraining motion. We call the set of roles instantiated in one system its \emph{Functional Role Profile}.

\begin{table}[t]
  \centering
  \small
  \begin{tabularx}{\columnwidth}{@{}p{0.30\columnwidth}X@{}}
    \toprule
    \rowcolor{repshade!28}
    \textbf{Role} & \textbf{Operational responsibility} \\
    \midrule
    \rowcolor{panoramarow}
    Specification & Establishes or revises the task that later behavior should satisfy. \\
    Embodied Representation & Makes task-relevant world or agent state available to later decisions or behavior. \\
    \rowcolor{panoramarow}
    Action Orchestration & Determines which available skills or agents should act and how their contributions are ordered or combined. \\
    Grounding Regulation & Uses new evidence to alter downstream computation. The alteration must reach later embodied behavior through a reported route. \\
    \rowcolor{panoramarow}
    Execution Coupling & Provides a language-derived quantity that is consumed in producing physical action. \\
    \bottomrule
  \end{tabularx}
  \caption{Operational definitions of five non-exclusive functional roles assigned to
  language-derived content.}
  \label{tab:roles}
\end{table}

Several boundaries keep these roles distinct. Specification concerns the task to be satisfied; Action Orchestration concerns how available capabilities are used to pursue that task. Embodied Representation makes state available to a later consumer, whereas Execution Coupling places a language-derived quantity on the path that produces physical action. Action Orchestration may remain open loop. Grounding Regulation begins only when new evidence changes downstream computation and that change reaches later embodied behavior.

Identifying a functional role does not show that the role is grounded. For each paper--role claim, we code five non-exclusive evidence operations.

\textbf{R: Route traceability.} A language-derived quantity, its downstream embodied consumer, and the route between them can be identified.

\textbf{T: Targeted behavioral test.} An intervention on the role-bearing quantity or interface produces a reported behavioral consequence.

\textbf{C: Embodied-constraint check.} An instance-matched embodied referent or constraint is used to check the role-bearing content during operation or evaluation.

\textbf{F: Closed-loop feedback.} An observed outcome triggers a revision that changes a later executable decision or attempt.

\textbf{I: Claim-relative isolation.} A matched comparison isolates the claimed role from its principal alternative explanation.

Isolation is relative to the stated claim. It need not control every source of complete-system competence. These operations are neither additional functional roles nor steps on a maturity scale. Appendix~\ref{app:evidence-codebook} gives the coding rules.

The warrant provided by an operation also depends on the source of the embodied check. Oracle or simulator state can test consistency with modeled constraints, but it cannot establish real-world robustness. Evidence from direct sensing or a physical outcome supports a different claim. For example, tactile contact can validate a represented physical property without showing that the agent recovers after failure. We refer to the source of a check as its \emph{evidence provenance}.

Corpus-level frequencies count distinct papers with at least one qualifying role claim. Assigning several roles to one paper does not create additional paper counts.

Section~\ref{sec:roles} uses the role definitions to compare embodied mechanisms. Section~\ref{sec:audit} applies the evidence operations to the claims reported in the reviewed literature.

%% file: sections/04_functional_roles.tex
\section{What Language Does in the Embodied Loop}
\label{sec:roles}

\input{figures/figure_role_landscape}

Figure~\ref{fig:role-landscape} shows that a single system may assign more than one
role to language. This section compares representative mechanisms for each role.
Appendix~\ref{app:core-panorama} provides the complete paper-level inventory underlying
this comparison.

\subsection{Specification: Committing the Agent to a Task}
\label{sec:specification}

\role{Specification} determines what later behavior should accomplish. Some systems
formalize an instruction against the observed scene. Others use available capabilities
when decomposing the task or revise the task during interaction.

Scene-linked formalization connects an instruction to entities and conditions that can
guide later behavior. \textit{ViLaIn} translates an instruction and an observed scene into PDDL
objects, an initial state, and a goal, making explicit what the planner should treat as
the task \citep{shirai2024vision}. \textit{Hi Robot} instead keeps the commitment revisable,
converting open-ended instructions and situated user interjections into updated commands
for a visuomotor policy \citep{shi2025hi}. These approaches differ in form: one
externalizes a structured task description, while the other maintains an instruction
that can change during interaction.

Capability information provides a further basis for specification. \textit{SMART-LLM} uses
descriptions of robot capabilities to determine how a task should be decomposed and what
capabilities its subtasks require \citep{kannan2024smart}. This establishes part of
the task commitment; assigning particular robots to those subtasks is a separate
orchestration responsibility discussed below. In these examples, language defines or
revises the task. The components that carry it out remain outside Specification.

\subsection{Embodied Representation: Making Embodied State Available}
\label{sec:representation}

\role{Embodied Representation} makes task-relevant state available to a later
consumer. It can describe the environment or actionable geometry. It can also preserve
state across interactions.

Situated representations connect linguistic distinctions to spatial or physical
properties. \textit{VLMaps} integrates open-vocabulary features into a persistent 3D map
used for navigation \citep{huang2023visual}. \textit{PhysObjects} represents physical
properties relevant to manipulation, such as material and fragility
\citep{gao2024physically}. \textit{Octopi} relates descriptions of physical properties to
tactile contact \citep{yu2024octopi}. Each representation makes a different kind of
embodied state available to downstream computation.

Some representations encode geometry in a form that can also shape motion. \textit{VoxPoser}
constructs dense spatial value fields, whereas \textit{ReKep} represents relations among tracked
keypoints \citep{huang2023voxposer,
huang2025rekep}. Their representational responsibility
lies in encoding actionable spatial state. A separate Execution Coupling responsibility
begins when those fields or relations are directly consumed by motion planning or
trajectory optimization.

Representation can also persist over time or describe the agent's emerging behavior.
\textit{Statler} maintains an explicit state for subsequent reasoning, while
\textit{DECKARD} updates an abstract world model through environment experience
\citep{yoneda2024statler,nottingham2023embodied}. \textit{ECoT} exposes an explicit
account of the current plan and task-relevant state, whereas \textit{RT-H} introduces a
language-like motion abstraction between task instructions and bodily commands
\citep{zawalski2025robotic,belkhale2024rthactionhierarchiesusing}. Persistent state
supports later reasoning. Action-oriented abstractions place the representation nearer
execution. Section~\ref{sec:audit} asks whether the represented state is accurate and
used downstream. It also asks whether embodied evidence can correct it.

\subsection{Action Orchestration: Organizing Embodied Capabilities}
\label{sec:orchestration}

\role{Action Orchestration} selects and arranges capabilities that already exist. It
does not supply their underlying motor competence.

Situated constraints can enter orchestration at different points. \textit{SayCan}
selects among learned robot skills by combining linguistic task relevance with a value
estimate conditioned on the current state \citep{ahn2022icanisay}. \textit{Grounded
Decoding} incorporates embodied constraints while generating a command
\citep{huang2023grounded}. \textit{SayPlan} instead checks a proposed action sequence
against an environment model \citep{rana2023sayplan}. The intervention point therefore
differs across the three systems. Each remains limited by the skills or environment
model available to it.

Orchestration can also allocate actors or routes. \textit{SMART-LLM} assigns
capability-compatible robots to subtasks after the required capabilities have been
specified, while \textit{CoNVOI} selects a contextually appropriate reference path among regions
that perception and motion-planning components have judged navigable
\citep{kannan2024smart,sathyamoorthy2024convoi}. In both cases, linguistic content
chooses among possibilities generated by other components. The selected robot or
planning stack remains responsible for execution.

\subsection{Grounding Regulation: Letting Evidence Change Behavior}
\label{sec:regulation}

\role{Grounding Regulation} uses new evidence to change later embodied behavior. The
change may occur before action, after a failed action, or across repeated attempts.

Before acting, \textit{KnowNo} estimates whether the instruction and current scene support a
sufficiently certain choice and requests clarification when they do not. During or after
action, \textit{Inner Monologue} returns observations and success or human feedback to the planner
so that the agent can retry or replan
\citep{ren2023robots,
huang2023inner}. \textit{KnowNo} therefore regulates whether action begins.
\textit{Inner Monologue} regulates the next decision after feedback contradicts an
earlier one.

Outcome evaluators occupy an adjacent but distinct position. \textit{AHA} identifies and explains
manipulation failures, while \textit{vision--language success detectors} estimate whether a
language-specified task has been completed
\citep{duan2024ahavisionlanguagemodeldetectingreasoning,du2023vision}. Producing such
a judgment does not by itself instantiate Grounding Regulation. The judgment must
change downstream computation through a reported route to later embodied behavior.

Persistent regulation is visible when corrections survive beyond one decision. \textit{DROC}
stores language corrections as retrievable rules or parameters that guide later
execution, while \textit{SAIL} places VLM feedback and retrieved trajectories inside
test-time search \citep{zha2024distilling,sato2026sailtesttimescalingincontext}. In both
systems, new evidence affects behavior beyond the current decision.

\subsection{Execution Coupling: Carrying Language into Physical Action}
\label{sec:coupling}

\role{Execution Coupling} places language-derived content on the action-production
path. The content may be exposed through an explicit interface or embedded within a
learned policy.

Explicit interfaces make the action route comparatively visible. \textit{Code as Policies}
generates programs that call perception and control primitives, while \textit{Language to
Rewards} supplies objectives optimized through model-predictive control
\citep{liang2023code,yu2023language}. \textit{VoxPoser} and \textit{ReKep} first encode actionable spatial
state, but they additionally instantiate Execution Coupling when their value fields or
relational constraints are consumed by motion planning or trajectory optimization
\citep{huang2023voxposer,
huang2025rekep}. In each system, language supplies a structure used in action
generation. Other components translate that structure into motion under the constraints
of the robot and its environment.

Learned policies place the linguistic interface inside action generation. \textit{RT-2}
represents robot actions as tokens within a vision--language model, \textit{CogACT} conditions a
diffusion action model on a visual--language representation, and \textit{RT-H} places a
language-like motion abstraction between task instructions and bodily commands
\citep{zitkovich2023rt,li2024cogactfoundationalvisionlanguageactionmodel,
belkhale2024rthactionhierarchiesusing}. Compared with explicit programs or constraints,
these learned interfaces bind semantic and action representations more tightly. This
also makes the linguistic contribution harder to inspect separately.
Section~\ref{sec:audit} evaluates whether the reported comparisons isolate that
contribution.

Because one mechanism may instantiate several roles, evidence must be assessed at the
paper--role level. The next section performs this assessment.

%% file: figures/figure_role_landscape.tex
\begin{figure*}[t]
\centering
\scriptsize
\setlength{\tabcolsep}{4pt}
\renewcommand{\arraystretch}{1.13}
\newcommand{\primarypaper}[2]{%
  \begingroup\setlength{\fboxsep}{0.9pt}\setlength{\fboxrule}{0.3pt}%
  \fcolorbox{black!22}{#1!52}{\strut\textit{#2}}\endgroup}
\begin{tabularx}{\textwidth}{@{}P{0.19\textwidth}Y@{}}
\toprule
\rowcolor{specshade!30}
\textbf{Specification}\par
Primary assignments: 15\newline Appears in full profile: 21 &
\primarypaper{specchip}{$\pi_0$} \primarypaper{specchip}{A2Nav} \primarypaper{specchip}{CLAIRIFY} \primarypaper{specchip}{CoELA} \primarypaper{specchip}{GR00T N1} \primarypaper{specchip}{Hi Robot} \primarypaper{specchip}{HRC Manipulation} \primarypaper{specchip}{Humanoid-LLA} \primarypaper{specchip}{Instruct2Act} \primarypaper{specchip}{LAGEA} \primarypaper{specchip}{Mobility VLA} \primarypaper{specchip}{OpenVLA} \primarypaper{specchip}{RoboBrain} \primarypaper{specchip}{SMART-LLM} \primarypaper{specchip}{ViLaIn} \\
\addlinespace[1.5pt]
\rowcolor{repshade!30}
\textbf{Embodied Representation}\par
Primary assignments: 56\newline Appears in full profile: 76 &
\primarypaper{repchip}{3D-VLA} \primarypaper{repchip}{3DLLM-Mem} \primarypaper{repchip}{Anticipation-VLA} \primarypaper{repchip}{ChatVLA} \primarypaper{repchip}{ChatVLA-2} \primarypaper{repchip}{CLIP-Nav} \primarypaper{repchip}{CogACT} \primarypaper{repchip}{CogVLA} \primarypaper{repchip}{CoNVOI} \primarypaper{repchip}{CoPa} \primarypaper{repchip}{CoT-VLA} \primarypaper{repchip}{Dejavu} \primarypaper{repchip}{DexGraspVLA} \primarypaper{repchip}{Distilled Feature Fields} \primarypaper{repchip}{ECoT} \primarypaper{repchip}{Embodied-RAG} \primarypaper{repchip}{EmbodiedGPT} \primarypaper{repchip}{EnerVerse} \primarypaper{repchip}{Explore until Confident} \primarypaper{repchip}{Fast-in-Slow} \primarypaper{repchip}{Fast-ThinkAct} \primarypaper{repchip}{FlowVLA} \primarypaper{repchip}{GPT-4V for Robotics} \primarypaper{repchip}{HAMSTER} \primarypaper{repchip}{JanusVLN} \primarypaper{repchip}{LEO} \primarypaper{repchip}{LGX} \primarypaper{repchip}{ManipLLM} \primarypaper{repchip}{Manipulate-Anything} \primarypaper{repchip}{MemER} \primarypaper{repchip}{MOKA} \primarypaper{repchip}{MOO} \primarypaper{repchip}{MultiPLY} \primarypaper{repchip}{NavCoT} \primarypaper{repchip}{NavGPT} \primarypaper{repchip}{NavGPT-2} \primarypaper{repchip}{Octopi} \primarypaper{repchip}{OmniManip} \primarypaper{repchip}{OneTwoVLA} \primarypaper{repchip}{PaLM-E} \primarypaper{repchip}{PhysObjects} \primarypaper{repchip}{PIVOT} \primarypaper{repchip}{RoboGround} \primarypaper{repchip}{RoboPoint} \primarypaper{repchip}{RT-H} \primarypaper{repchip}{SayPlan} \primarypaper{repchip}{Scene-LLM} \primarypaper{repchip}{SpatialVLA} \primarypaper{repchip}{Statler} \primarypaper{repchip}{TaPA} \primarypaper{repchip}{ThinkAct} \primarypaper{repchip}{TLA} \primarypaper{repchip}{UP-VLA} \primarypaper{repchip}{Visual Language Maps} \primarypaper{repchip}{VoxPoser} \primarypaper{repchip}{VTLA} \\
\addlinespace[1.5pt]
\rowcolor{orchshade!30}
\textbf{Action Orchestration}\par
Primary assignments: 12\newline Appears in full profile: 68 &
\primarypaper{orchchip}{DECKARD} \primarypaper{orchchip}{DEPS} \primarypaper{orchchip}{Grounded Decoding} \primarypaper{orchchip}{LLM-DP} \primarypaper{orchchip}{LLM-Planner} \primarypaper{orchchip}{Octopus} \primarypaper{orchchip}{ProgPrompt} \primarypaper{orchchip}{ReplanVLM} \primarypaper{orchchip}{SayCan} \primarypaper{orchchip}{SEEA-R1} \primarypaper{orchchip}{Unified Agent} \primarypaper{orchchip}{Zero-Shot Planners} \\
\addlinespace[1.5pt]
\rowcolor{regshade!30}
\textbf{Grounding Regulation}\par
Primary assignments: 18\newline Appears in full profile: 52 &
\primarypaper{regchip}{AHA} \primarypaper{regchip}{APO} \primarypaper{regchip}{DROC} \primarypaper{regchip}{ELITE} \primarypaper{regchip}{EmbodiSkill} \primarypaper{regchip}{Evolvable Embodied Agent} \primarypaper{regchip}{GRAPE} \primarypaper{regchip}{Hume} \primarypaper{regchip}{Inner Monologue} \primarypaper{regchip}{KnowNo} \primarypaper{regchip}{Language to Rewards} \primarypaper{regchip}{Large Reward Models} \primarypaper{regchip}{Reflective Planning} \primarypaper{regchip}{RoboFuME} \primarypaper{regchip}{SAIL} \primarypaper{regchip}{Text2Reward} \primarypaper{regchip}{VLA-RL} \primarypaper{regchip}{VLM-Social-Nav} \\
\addlinespace[1.5pt]
\rowcolor{couplingshade!30}
\textbf{Execution Coupling}\par
Primary assignments: 4\newline Appears in full profile: 62 &
\primarypaper{couplingchip}{Code as Policies} \primarypaper{couplingchip}{DexVLA} \primarypaper{couplingchip}{ReKep} \primarypaper{couplingchip}{RT-2} \\
\bottomrule
\end{tabularx}
\caption{Functional role profiles in the reviewed literature. For each role, the
figure reports the number of papers with that primary assignment and the number in
which the role appears anywhere in the full profile. The paper labels show the primary
assignments. Each paper receives one primary assignment but may instantiate multiple
roles. Appendix~\ref{app:core-panorama} provides the corresponding paper-level
inventory.}
\label{fig:role-landscape}
\end{figure*}

%% file: sections/05_evidence_audit.tex
\section{What the Evidence Supports}
\label{sec:audit}

This section evaluates how the reported evidence supports claims about each functional
role. We focus on four recurrent mismatches between an observation and the claim it is
used to support. The analysis identifies missing embodied checks or unresolved
alternatives. It does not rank systems.

\subsection{Evidence Operations across the Reviewed Corpus}

\input{tables/table4_evidence_coverage}

All 105 papers satisfy R because a traceable route from language-derived content to a
downstream consumer is an inclusion requirement. T appears in 97 papers (92.4\%), and C
appears in 76 (72.4\%). I appears in 85 papers (81.0\%). An I-positive comparison
supports attribution relative to the principal alternative; it does not explain
complete-system competence. F is less common, appearing in 30 papers (28.6\%). F
requires an observed outcome to trigger a revision that changes a later executable
attempt. It does not by itself establish successful recovery.

These counts are marginal frequencies of reported evidence operations. They do not form
a universal grounding test or a ranking of systems. Each operation must be interpreted
relative to the claimed role, and role prevalence does not determine grounding strength.
Appendix~\ref{app:evidence-profiles} reports the joint profiles underlying these
frequencies.

\subsection{Inspectable Language Is Not Necessarily Correct or Used}

An explicit language-derived artifact is easier to inspect. It reveals what content is
available to a downstream consumer, but visibility alone does not show that the content
matches the embodied world or changes behavior. \textit{ReKep} evaluates its keypoint
constraints against tracked geometry and reports their use in motion optimization under
disturbances. \textit{Statler}'s explicit state supports later reasoning, but it may be
updated from the semantics of an attempted action without an independent observation
that the action succeeded \citep{huang2025rekep,yoneda2024statler}. \textit{DECKARD}
makes the contrast explicit: an LLM-proposed transition is verified only after the agent
reaches it through environment experience, allowing observation to correct the abstract
world model \citep{nottingham2023embodied}.

Correctness and causal use require different evidence. \textit{ViLaIn} can produce a
parsable and solvable problem description without establishing that its objects and
initial predicates match the observed scene \citep{shirai2024vision}. \textit{ECoT}
exposes instance-specific boxes and robot-state fields and permits human correction,
providing evidence about both content and downstream use \citep{zawalski2025robotic}.
These interventions do not show that the trace explains complete-system success. A
representation can be formally valid without matching the observed scene, and it can
match the scene without being used in behavior. An explicit artifact makes these
distinctions testable; readability alone does not resolve them.

\subsection{Internal Revision Is Not Embodied Closure or Successful Recovery}

A feedback signal can be valid without closing a behavioral loop, and a loop can consume
feedback whose validity has not been established. \textit{Vision--language success
detectors} are primarily evaluated post hoc. \textit{AHA} also detects and explains
manipulation failures, but additionally reports downstream integrations in which its
feedback changes later behavior
\citep{du2023vision,duan2024ahavisionlanguagemodeldetectingreasoning}. Detection results
bear on evaluator quality. Grounding Regulation begins only when the verdict changes
downstream computation and reaches later embodied behavior. \textit{Inner Monologue}
exposes this route by returning observations and success or human feedback to a planner
that can retry or replan \citep{huang2023inner}.

Feedback consumption alone does not establish embodied closure. An internal artifact may
change while execution remains unaffected. Closure requires the revision to alter a
later executable attempt. Successful recovery further requires evidence that the outcome
improved. \textit{ViLaIn} and \textit{SayPlan} use formal or simulated errors to
regenerate content. \textit{DROC} stores human corrections for later execution, while
\textit{SAIL} uses rollout feedback to modify test-time search
\citep{shirai2024vision,rana2023sayplan,zha2024distilling,sato2026sailtesttimescalingincontext}.
A trace ending at internal revision supports only a revision claim. A changed executable
attempt supports closure. Recovery additionally requires an improved outcome.

\subsection{System Success Alone Does Not Isolate Language}

Task success evaluates the complete embodied system but rarely identifies which
component produced the gain. \textit{RT-2} demonstrates broad language-conditioned robot
behavior, while several potential sources of that performance remain coupled
\citep{zitkovich2023rt}. \textit{CogACT} exposes one such alternative directly: with its
vision--language foundation held fixed, changing the diffusion action module
substantially changes success \citep{li2024cogactfoundationalvisionlanguageactionmodel}.
This is valid system-level evidence, but it cannot be treated as evidence of stronger
language grounding by default.

Claim-relative isolation can nevertheless narrow the explanation. \textit{RT-H}
compares one-hot, clustered, hierarchical, and flat motion representations, while
\textit{ECoT} permits intervention on an embodied trace
\citep{belkhale2024rthactionhierarchiesusing,zawalski2025robotic}. These comparisons
isolate the targeted representation or trace against a specified alternative. They do
not account for every source of complete-system competence.

\subsection{Action Proximity Does Not Determine Grounding Strength}

Language-derived quantities can enter action production through explicit interfaces or
learned representations. Their architectural distance from motor output varies, but none
produces motion alone. \textit{Code as Policies} relies on the available perception and
control primitives. \textit{VoxPoser} and \textit{ReKep} rely on tracked state and
downstream optimization. Learned policies rely on action models trained with robot data
\citep{liang2023code,huang2023voxposer,huang2025rekep,zitkovich2023rt}. Execution
outcomes test this complete responsibility chain. They do not isolate the
language-derived quantity.

Architectural coupling alone does not provide stronger evidence. \textit{SayCan}
separates linguistic relevance from state-conditioned feasibility, making the two
contributions comparatively inspectable even though the rest of the system remains
important \citep{ahn2022icanisay}. Direct conditioning of an action model may place
linguistic content closer to motor output while making its contribution harder to
perturb separately. \textit{CoNVOI} distinguishes contextual route preference from
physical traversability. Language informs the former, whereas lidar and motion planning
provide evidence for the latter \citep{sathyamoorthy2024convoi}. Architectural proximity
to action is therefore not evidence of stronger grounding. The relevant question is
whether the reported checks bear on the claimed contribution and address its principal
alternative.

%% file: tables/table4_evidence_coverage.tex
\begin{table}[!t]
\centering
\small
\setlength{\tabcolsep}{8pt}
\renewcommand{\arraystretch}{1.10}
\begin{tabular}{@{}lr@{}}
\toprule
\textbf{Evidence operation} & \textbf{Papers (\%)} \\
\midrule
\rowcolor{panoramarow}
Route traceability (R) & 105 (100.0\%) \\
Targeted behavioral test (T) & 97 (92.4\%) \\
\rowcolor{panoramarow}
Embodied-constraint check (C) & 76 (72.4\%) \\
Closed-loop feedback (F) & 30 (28.6\%) \\
\rowcolor{panoramarow}
Claim-relative isolation (I) & 85 (81.0\%) \\
\bottomrule
\end{tabular}
\caption{Evidence operations reported in the reviewed corpus. Counts aggregate
paper--role judgments to distinct papers ($n=105$), and the operations are
non-exclusive. R reaches 100\% because route traceability is an inclusion requirement.
The frequencies do not form a score or maturity scale. I evaluates a claimed role
relative to its principal alternative; it does not decompose complete-system
competence.}
\label{tab:evidence-coverage}
\end{table}

%% file: sections/06_implications.tex
\section{Implications for Evaluation and Reporting}
\label{sec:implications}

Evaluation should begin with the responsibility assigned to language rather than with an
end-to-end score. Once that responsibility is stated, the reported evidence can be
judged against a specific claim about what language-derived content changes in the
embodied system.

\subsection{State the Responsibility Chain}

A report should identify the language-derived quantity under study and the downstream
component that consumes it. It should then explain how a change in that quantity is
expected to reach embodied behavior. When one mechanism instantiates several roles,
the claim and its supporting evidence should be stated separately for each role. Route
traceability makes the responsibility chain inspectable, but it does not establish that
the content is correct or causally used. The report should therefore also state what
behavioral consequence is expected if the role-bearing quantity is altered.

\subsection{Design Tests around the Claimed Role}

Task outcomes remain informative, but role-specific claims require tests directed at
the relevant responsibility. The appropriate intervention depends on what language is
claimed to do.

\textbf{Specification.} An evaluation can vary the task formulation while preserving
the observed scene and the capabilities available to the agent. The resulting comparison
tests whether the language-derived task commitment changes later behavior.

\textbf{Embodied Representation.} The represented content can be checked against an
instance-matched observation before it reaches a downstream consumer. An additional
intervention can test whether changing that representation affects a later decision or
action.

\textbf{Action Orchestration.} A comparison can hold the available capability set fixed
while changing how capabilities are selected, ordered, or assigned. This targets the
orchestration mechanism without attributing the underlying motor competence to language.

\textbf{Grounding Regulation.} The evaluation should trace new evidence to a change in
downstream computation and then to a later executable attempt. A feedback signal or an
internal revision alone does not establish that behavior was regulated.

\textbf{Execution Coupling.} Where the architecture permits, the action-bearing
language-derived quantity or interface can be perturbed while the rest of the action
stack is preserved. This tests whether the claimed coupling contributes to physical
action without treating the complete controller as a language effect.

A role-specific comparison need not isolate every source of system competence. It
should, however, address the principal alternative to the role-specific claim and state
which other explanations remain unresolved.

\subsection{Distinguish Revision, Closure, and Recovery}

An internal change supports a revision claim. If that change reaches a later executable
attempt, the evidence supports embodied closure. Recovery additionally requires an
observed improvement or successful outcome. Reports should identify the last point in
this chain that was directly observed instead of describing every feedback-enabled
system as closed loop or self-correcting.

The same discipline applies to attribution. A matched comparison may isolate the
contribution of a role-bearing quantity relative to a specified alternative. It does not
show that language explains complete-system competence. Authors should name the
alternative addressed by the comparison and limit the attribution accordingly.

\subsection{Preserve the Scope of the Evidence}

Evidence provenance determines what a check can establish. Oracle or simulator state
can test consistency with modeled constraints. Direct sensing can test correspondence
with the measured environment. A learned success detector supports a behavioral claim
only to the extent that the detector has been validated for the setting in which it is
used. A physical outcome can establish whether the task was completed, but it does not
by itself identify which component produced that outcome. Reports should therefore name
the source of each embodied check and the downstream process that consumes it.

The evidence operations should be reported as a role-specific R/T/C/F/I profile rather
than collapsed into a single grounding score. A missing operation does not automatically
show that a system is ungrounded. It limits the claims warranted by the reported
evaluation. Appendix~\ref{app:core-panorama} provides the paper inventory by primary
role and evidence profile.

\begin{focusbox}[\faCheckCircle\ Claim--Evidence Contract]
A claim--evidence contract requires the reported evidence to connect a stated language
role to later embodied behavior. The provenance of that evidence must be appropriate to
the claim, and the attribution can extend only as far as the principal alternative that
the evaluation rules out. When the responsibility chain ends, the claim must end there
as well.
\end{focusbox}

%% file: sections/07_conclusion.tex
\section{Conclusion}
\label{sec:conclusion}

By separating five non-exclusive language responsibilities from the evidence used to
warrant them, this review makes modular planners, embodied representations, feedback
systems, and integrated policies comparable without reducing them to an architecture or
system stage. Across these roles, matched comparisons often support the contribution of
a specific language-mediated quantity or interface, but such local attribution does not
by itself explain complete-system competence. Evidence that observed embodied outcomes
actually revise a later executable attempt remains substantially less common. The
resulting framework turns language grounding from a broad system label into a set of
testable, role-specific questions for interpreting current results and designing more
informative embodied evaluations.

%% file: sections/08_limitations.tex
\section*{Limitations}

This review evaluates evidence reported in the original papers rather than independently
reproducing each system. Its conclusions therefore concern what the published experiments
support, not whether an unreported grounding mechanism is absent. The five functional
roles are non-exclusive, and some responsibility chains permit more than one reasonable
interpretation. We reduce this ambiguity by consistently identifying the linguistic
content, its consumer, the embodied check, and the principal alternative explanation,
while acknowledging that the analysis still involves expert judgment. Evidence
provenance also bounds the conclusions: results based on labels, symbolic state,
simulation, direct sensing, and physical execution support different claims, and
simulation alone does not establish real-world robustness.

The corpus was constructed for traceable mechanism coverage rather than exhaustive
recall or field-wide prevalence estimation. Quantitative results are reported only where
paper--role claims were evaluated under a uniform rule and aggregated by distinct paper.
One author led the coding, with key and borderline cases discussed by the author team;
we therefore do not claim rater-independent ground truth or inter-rater reliability.
Claim-relative isolation also does not decompose complete-system competence. The
paper-level examples in Sections~\ref{sec:roles} and~\ref{sec:audit} illustrate
contrasts identified during corpus coding; they do not determine the taxonomy,
eligibility decisions, or corpus-level counts. Because recent
embodied-agent mechanisms often first appear as preprints, publication status was not
used as an exclusion criterion, and all conclusions remain conditional on the reported
paper versions.

%% file: sections/09_ethical_considerations.tex
\section*{Ethical Considerations}

This survey introduces no new human-subject data, robot experiments, or model training.
Its principal ethical risk lies in interpretation: citation prominence may reflect
institutional visibility, access to well-resourced robotic platforms, or English-language
documentation rather than scientific importance. Representative examples should
therefore not be read as a ranking of papers or research groups.

The audit is likewise not a scalar ranking or a safety assessment. Explicit intermediate
representations, calibrated uncertainty, and language-based failure explanations can
improve inspectability without guaranteeing physical feasibility, collision avoidance,
or safe execution. Claims about deploying embodied agents must remain accountable to
hardware safeguards, environment assumptions, human oversight, and real-world
consequences beyond language-conditioned benchmark success.

%% file: appendices/appendices.tex
\section{Review Protocol and Corpus Construction}
\label{app:review-protocol}

\subsection{Scope and Discovery}

This appendix documents paper discovery, scope screening, and evidence coding.
Eligibility followed the operational scope in Section~\ref{sec:scope}: a pretrained
language-aligned foundation model had to materially affect an embodied
perception--decision--action or learning loop. Search terms identified candidates but
did not determine their functional roles or evidential strength. The resulting corpus
supports traceable mechanism comparison rather than exhaustive recall or estimates of
field-wide prevalence.

\begin{table}[!t]
\centering
\scriptsize
\setlength{\tabcolsep}{3pt}
\renewcommand{\arraystretch}{1.04}
\begin{tabularx}{\columnwidth}{@{}P{0.48\columnwidth}Y@{}}
\toprule
\rowcolor{repshade!28}
\multicolumn{2}{c}{\textbf{Embodied-domain phrases}} \\
\rowcolor{repshade!14}
\textbf{Family} & \textbf{Search phrase} \\
\midrule
\rowcolor{panoramarow}
Vision--language--action & ``vision language action'' \\
Embodied agents & ``embodied agent'' \\
\rowcolor{panoramarow}
Robotic manipulation & ``robot manipulation'' \\
Robot navigation & ``robot navigation'' \\
\rowcolor{panoramarow}
Embodied task planning & ``robot task planning'' \\
Robot policies/action & ``robot policy'' \\
\rowcolor{panoramarow}
Generalist robots & ``generalist robot'' \\
\addlinespace[2pt]
\rowcolor{repshade!28}
\multicolumn{2}{c}{\textbf{Foundation-model phrases}} \\
\rowcolor{repshade!14}
\textbf{Family} & \textbf{Search phrase} \\
\midrule
\rowcolor{panoramarow}
Large language models & ``large language model'' \\
Vision--language models & ``vision language model'' \\
\rowcolor{panoramarow}
Multimodal large language models & ``multimodal large language model'' \\
Robotics foundation models & ``robot foundation model'' \\
\bottomrule
\end{tabularx}
\caption{Search phrases used for database retrieval. Each query combined one
embodied-domain phrase with one foundation-model phrase. Ambiguous acronyms were
excluded as standalone search terms.}
\label{tab:search-matrix}
\end{table}

\begin{table}[!t]
\centering
\scriptsize
\setlength{\tabcolsep}{3pt}
\renewcommand{\arraystretch}{1.08}
\begin{tabularx}{\columnwidth}{@{}P{0.31\columnwidth}rY@{}}
\toprule
\rowcolor{repshade!28}
\textbf{Stage} & \textbf{$n$} & \textbf{Description} \\
\midrule
\rowcolor{panoramarow}
Database search & 1,478 & Unique records after deduplication \\
Database subset screened & 148 & Highest-cited 10\% of retrieved records \\
\rowcolor{panoramarow}
Additional discovery & 70 & Papers retained from reading and citation tracing \\
Merged set & 199 & Records after merging and version deduplication \\
\rowcolor{panoramarow}
Initial scope screen & 151 & Technical papers remaining after clear exclusions \\
Final corpus & 105 & Papers included in role and evidence coding \\
\bottomrule
\end{tabularx}
\caption{Summary of corpus construction. Counts describe discovery and screening, not
levels of evidence.}
\label{tab:corpus-flow}
\end{table}

We searched titles and abstracts in the Semantic Scholar Academic Graph without
restricting venue, publication status, open-access availability, or citation count. The
search returned more records than could be screened manually, so we screened the
highest-cited 10\% as a prioritization sample. The cutoff was a practical screening
limit, not an eligibility criterion or a proxy for evidential quality, and it favors
older work.

\begin{table*}[!t]
\centering
\scriptsize
\setlength{\tabcolsep}{3.2pt}
\renewcommand{\arraystretch}{1.12}
\begin{tabularx}{\textwidth}{@{}P{0.075\textwidth}P{0.29\textwidth}P{0.29\textwidth}Y@{}}
\toprule
\rowcolor{repshade!28}
\textbf{Operation} & \textbf{Qualifying rule} &
\textbf{Evidence that does not qualify} & \textbf{Interpretive boundary} \\
\midrule
\rowcolor{panoramarow}
\textbf{R: Route traceability} &
A language-derived quantity, a downstream embodied consumer, and a traceable route
between them are all identifiable. &
Language appears only in the task description, user interface, training corpus, or
system narrative, with no reported route to embodied computation. &
R establishes functional inclusion and mechanism traceability, not that the quantity is
correct, necessary, physically checked, or independently causal. \\

\textbf{T: Targeted behavioral test} &
The specific role-bearing quantity or interface is removed, replaced, perturbed,
corrupted, or controlled, and a behavioral consequence is reported. &
A full-system comparison, model-scale or data-size change, generic component ablation,
or alternative controller that changes several non-target factors. &
The intervention must target the claimed language responsibility. T can be positive even
when the comparison does not isolate that responsibility from its principal alternative. \\

\rowcolor{panoramarow}
\textbf{C: Embodied-constraint check} &
An instance-matched embodied referent or constraint evaluates, gates, rejects, revises,
or selects the role content during operation or evaluation. Valid provenance includes
sensing, simulator or solver state, annotations, contact, and real-robot outcomes. &
Merely including a simulator, planner, perception module, benchmark, or robot; checking
only textual plausibility; or reporting task success without showing what embodied
referent checks the role content. &
C warrants contact with a specified embodied constraint. Its strength remains bounded
by provenance: oracle or simulated checks do not by themselves establish sensing
robustness or real-world validity. \\

\textbf{F: Closed-loop feedback} &
An observed outcome triggers a revision that changes a subsequent executable decision
or physical attempt. &
An offline evaluator, feedback that is available but unused, recurrent observation
without revision, text-only reflection, or restarting the same attempt unchanged. &
F concerns closure of the observation--revision--action loop. It does not require the
revised attempt to succeed, and success alone does not demonstrate that such a loop
occurred. \\

\rowcolor{panoramarow}
\textbf{I: Claim-relative isolation} &
A matched comparison isolates the role-bearing quantity or interface relative to the
principal alternative explanation for the claim while holding major non-target factors
fixed. &
An unmatched baseline, complete-system leaderboard gain, or comparison that
simultaneously changes model, data, perception, controller, and language interface. &
I supports attribution of the stated role claim, not a causal decomposition of all
complete-system competence. It is stricter than observing a targeted effect, so T need
not imply I. \\
\bottomrule
\end{tabularx}
\caption{Operational codebook for the five evidence operations. Rules apply to a
paper--role claim. Items in the exclusion column are common false positives, not claims
that the listed evidence has no other scientific value.}
\label{tab:evidence-operation-codebook}
\end{table*}

We supplemented the database search with papers found through exploratory reading and
citation tracing. This additional search was intended to recover recent work and
mechanisms described with terminology not captured by the database queries. It retained
63 technical papers and seven benchmark or resource papers after the same scope criteria
were applied. The two sets were then merged and deduplicated before final screening.

The database search covered papers published from January 2021 through July 2026.
Earlier papers found through reading or citation tracing were included only when they
satisfied the same scope criteria. Table~\ref{tab:search-matrix} lists the two groups of
phrases combined in the database queries. Ambiguous abbreviations such as VLA, VLN,
LLM, VLM, and MLLM were not used as standalone terms, and semantically redundant
pairings were omitted.

\subsection{Corpus Construction}

The initial title-and-abstract screen removed clear scope violations, including
autonomous driving, GUI or computer agents, environment generation, text-only agents,
and systems in which language did not materially participate in an embodied loop. Of
the 165 technical papers in the merged set, 14 were removed at this stage. The remaining
151 were assessed for the functional roles assigned to language and the evidence
reported for those roles; 105 met the criteria for final role and evidence coding.
Papers outside the final corpus were not included in corpus-level counts.

The database search and supplementary discovery process have different selection
biases. Their union should therefore be interpreted as the reviewed corpus, not as an
estimate of field prevalence or retrieval recall.

\subsection{Evidence Coding}
\label{app:evidence-codebook}

The coding unit was a \emph{paper--role claim}. Each paper in the final corpus received
a non-exclusive functional-role profile, and evidence operations were coded separately
for each qualifying role claim. When the reported mechanism or evidence was unclear,
the relevant method and experiment sections were consulted before aggregation. Corpus
frequencies count a distinct paper once when at least one qualifying role claim
satisfies an operation. An operation assigned to one role does not automatically
transfer to the paper's other roles. The operations are non-exclusive forms of warrant
rather than a score or maturity sequence. Route traceability (R) was the common
inclusion baseline.

\begin{table*}[!t]
\centering
\scriptsize
\setlength{\tabcolsep}{3.2pt}
\renewcommand{\arraystretch}{1.12}
\begin{tabularx}{\textwidth}{@{}P{0.18\textwidth}P{0.22\textwidth}P{0.27\textwidth}Y@{}}
\toprule
\rowcolor{repshade!28}
\textbf{Review and scope} & \textbf{Primary organizing unit} &
\textbf{Treatment of language} & \textbf{Relation to this review} \\
\midrule

\citet{wang2024largelanguagemodelsrobotics}: large language and multimodal models in robotics &
Robot capabilities, including planning, manipulation, and reasoning &
Language supports instruction understanding and reasoning within a broader model
capability stack. &
Physical evaluation is surveyed, but language-specific responsibility and attribution
are not the organizing units. \\

\rowcolor{panoramarow}
\citet{ma2026survey}: vision--language--action models for embodied AI &
Model components, policies, planners, datasets, and benchmarks &
Language is one modality connecting instructions, perception, and action. &
The survey maps the VLA stack; our audit separates the claimed role of language from
evidence supplied by perception, action learning, data, or control. \\

\citet{liang2025largemodelempoweredembodied}: large-model-empowered embodied AI &
Hierarchical and end-to-end decision making, embodied learning, and world models &
Language participates in planning, feedback, and learning across system paradigms. &
Our unit is a role-specific chain of influence that cuts across these paradigms. \\

\rowcolor{panoramarow}
\citet{chen2025exploring}: embodied multimodal large models &
Perception, navigation, interaction, simulation, and supporting resources &
Language is integrated with multimodal sensing and reasoning. &
Our review asks what linguistic content changes and what embodied evidence warrants
that change. \\

\citet{kawaharazuka2025vision}: VLA robotics toward real-world use &
Architectures, modalities, learning strategies, platforms, and data &
Language conditions a deployment-oriented perception--action stack. &
Our audit keeps downstream perception, action modeling, and control distinct from
language-specific attribution. \\

\rowcolor{panoramarow}
\citet{shao2025largevlmbasedvisionlanguageactionmodels}: large-VLM-based VLA manipulation &
Monolithic and hierarchical architectures, learning paradigms, and benchmarks &
Language appears in semantic conditioning, planning, and intermediate representations. &
Our functional roles compare these mechanisms across architectures and pair each role
with a corresponding embodied test. \\

\citet{chen2026semanticlifecycleembodiedai}: foundation-model-driven semantic knowledge &
Acquisition, representation, and storage of semantic knowledge across an embodied
lifecycle &
Language contributes to the continuity of semantic information. &
Our audit asks whether a particular language-mediated state changes behavior and remains
answerable to embodied evidence. \\

\rowcolor{TakeawayBg}
\textbf{This review}: language grounding across foundation-model-enabled modular and
end-to-end embodied agents &
Non-exclusive functional responsibility and its evidence &
Language is assigned one or more of five non-exclusive functional roles within the
embodied loop. &
For each role, we trace the responsibility chain and evaluate the reported evidence
against the claim. \\

\bottomrule
\end{tabularx}
\caption{Comparison with representative surveys. Differences in organizing question
describe analytical scope rather than coverage or quality.}
\label{tab:survey-comparison}
\end{table*}

One author performed the initial coding and led the evidence recoding. Before the counts
were finalized, the author team reviewed cases marked as ambiguous by the initial coder
and the paper-level contrasts discussed in the main text. This review resolved coding
decisions but was not an independent second annotation. For each embodied-constraint
judgment, we recorded whether the instance-matched check came from sensing, simulator
or solver state, reference annotations, contact, or real-robot outcomes.
Claim-relative isolation required a matched comparison against the principal
alternative, not decomposition of every contributor to complete-system competence. The
counts therefore describe an author-led analysis of this corpus rather than field
prevalence or rater-independent ground truth.

\section{Comparison with Related Surveys}
\label{app:survey-comparison}

Existing surveys offer complementary maps of foundation-model-enabled embodied
intelligence, but they differ in what they make the primary unit of comparison. We
compare their organizing questions rather than their overall coverage or quality. The
purpose is to locate the analytical space addressed by this review, not to argue that
role--evidence analysis replaces capability-, architecture-, or lifecycle-oriented
surveys.

Table~\ref{tab:survey-comparison} compares representative surveys by organizing unit and
treatment of language. The final column states how each perspective relates to the
role--evidence question developed in this review.

\makeatletter
\setlength{\@dblfptop}{0pt}
\makeatother
\setcounter{dbltopnumber}{3}
\renewcommand{\dbltopfraction}{0.97}
\renewcommand{\dblfloatpagefraction}{0.5}

\section{Paper Inventory by Primary Role and Evidence Profile}
\label{app:core-panorama}

This appendix provides the paper-level inventory underlying the corpus statistics in
Sections~\ref{sec:roles} and~\ref{sec:audit}. Each of the 105 reviewed papers appears
once under its coded primary role and exact R/T/C/F/I profile. Supporting roles remain
non-exclusive and are summarized at the corpus level in
Figure~\ref{fig:role-landscape}. Profile order follows corpus frequency and is not a
maturity or quality ranking. Each table cites the included papers in full.

Publication status was verified for all 105 papers using the latest bibliographic
records available during final corpus checking. Papers marked with
\textsuperscript{$\dagger$} were available only as preprints; the remaining 68 papers
had a verified conference or journal publication record. This annotation reports
publication status only and is not used as a proxy for evidential quality.

\newenvironment{coreinventorytable}[2]{%
\begin{table*}[t]
\def\coreinventorycaption{#1}%
\def\coreinventorylabel{#2}%
\centering
\scriptsize
\setlength{\tabcolsep}{3.2pt}
\renewcommand{\arraystretch}{1.10}
\begin{tabular}{@{}P{0.13\textwidth}P{0.05\textwidth}P{0.76\textwidth}@{}}
\toprule
\rowcolor{repshade!28}
\textbf{Exact profile} & \textbf{$n$} & \textbf{Papers with this primary role} \\
\midrule
}{%
\bottomrule
\end{tabular}
\caption{\coreinventorycaption}\label{\coreinventorylabel}
\end{table*}
}

\begin{coreinventorytable}{Paper inventory for Specification.}{tab:core-inventory-specification}
\textbf{R/T/C/I} & 5 & Humanoid-LLA\textsuperscript{$\dagger$} \citep{liu2026commandinghumanoidfreeformlanguage}; Instruct2Act\textsuperscript{$\dagger$} \citep{huang2023instruct2actmappingmultimodalityinstructions}; LAGEA\textsuperscript{$\dagger$} \citep{chowdhury2026lagealanguageguidedembodied}; Mobility VLA\textsuperscript{$\dagger$} \citep{chiang2024mobilityvlamultimodalinstruction}; SMART-LLM \citep{kannan2024smart} \\
\rowcolor{panoramarow}\textbf{R/T/C/F/I} & 3 & CLAIRIFY\textsuperscript{$\dagger$} \citep{skreta2023errorsusefulpromptsinstruction}; Hi Robot \citep{shi2025hi}; ViLaIn \citep{shirai2024vision} \\
\textbf{R/T/I} & 2 & A2Nav\textsuperscript{$\dagger$} \citep{chen2023a2navactionawarezeroshotrobot}; CoELA \citep{zhang2024building} \\
\rowcolor{panoramarow}\textbf{R/T} & 1 & $\pi_0$\textsuperscript{$\dagger$} \citep{black2026pi0visionlanguageactionflowmodel} \\
\textbf{R} & 2 & GR00T N1\textsuperscript{$\dagger$} \citep{nvidia2025gr00tn1openfoundation}; OpenVLA \citep{kim2025openvla} \\
\rowcolor{panoramarow}\textbf{R/T/C} & 1 & HRC Manipulation \citep{liu2024enhancing} \\
\textbf{R/C} & 1 & RoboBrain \citep{ji2025robobrain} \\
\end{coreinventorytable}

\begin{coreinventorytable}{Paper inventory for Embodied Representation.}{tab:core-inventory-embodied-representation}
\textbf{R/T/C/I} & 22 & CLIP-Nav \citep{dorbala2022clip}; CoNVOI \citep{sathyamoorthy2024convoi}; CoPa \citep{huang2024copa}; Embodied-RAG\textsuperscript{$\dagger$} \citep{xie2025embodiedraggeneralnonparametricembodied}; Explore until Confident\textsuperscript{$\dagger$} \citep{ren2024exploreconfidentefficientexploration}; GPT-4V for Robotics \citep{wake2024gpt} \\
\rowcolor{panoramarow}\textbf{R/T/C/I (cont.)} &  & LGX \citep{dorbala2023can}; ManipLLM \citep{li2024manipllm}; MOKA \citep{fang2024moka}; MOO \citep{stone2023moo}; MultiPLY \citep{hong2024multiply}; NavCoT \citep{lin2025navcot} \\
\textbf{R/T/C/I (cont.)} &  & NavGPT \citep{zhou2024navgpt}; Octopi \citep{yu2024octopi}; OmniManip \citep{pan2025omnimanip}; PIVOT \citep{nasiriany2024pivot}; RoboGround \citep{huang2025roboground}; Scene-LLM\textsuperscript{$\dagger$} \citep{fu2024scenellmextendinglanguagemodel} \\
\rowcolor{panoramarow}\textbf{R/T/C/I (cont.)} &  & TaPA\textsuperscript{$\dagger$} \citep{wu2023embodiedtaskplanninglarge}; TLA\textsuperscript{$\dagger$} \citep{hao2025tlatactilelanguageactionmodelcontactrich}; Visual Language Maps \citep{huang2023visual}; VTLA \citep{zhang2026vtla} \\
\textbf{R/T/C/F/I} & 5 & Dejavu\textsuperscript{$\dagger$} \citep{wu2026dejavuexperiencefeedbacklearning}; ECoT \citep{zawalski2025robotic}; RT-H\textsuperscript{$\dagger$} \citep{belkhale2024rthactionhierarchiesusing}; SayPlan \citep{rana2023sayplan}; ThinkAct \citep{huang2026thinkact} \\
\rowcolor{panoramarow}\textbf{R/T/I} & 17 & 3D-VLA \citep{zhen20243d}; 3DLLM-Mem \citep{hu20263dllm}; Anticipation-VLA\textsuperscript{$\dagger$} \citep{zhang2026anticipationvlasolvinglonghorizonembodied}; ChatVLA-2\textsuperscript{$\dagger$} \citep{zhou2025chatvla2visionlanguageactionmodelopenworld}; CogVLA \citep{li2026cogvla}; CoT-VLA \citep{zhao2025cot} \\
\textbf{R/T/I (cont.)} &  & EmbodiedGPT \citep{mu2023embodiedgpt}; Fast-in-Slow \citep{chen2026fast}; Fast-ThinkAct\textsuperscript{$\dagger$} \citep{huang2026fastthinkactefficientvisionlanguageactionreasoning}; FlowVLA\textsuperscript{$\dagger$} \citep{zhong2025flowvlavisualchainthoughtbased}; HAMSTER \citep{li2025hamster}; JanusVLN\textsuperscript{$\dagger$} \citep{zeng2026janusvlndecouplingsemanticsspatiality} \\
\rowcolor{panoramarow}\textbf{R/T/I (cont.)} &  & MemER\textsuperscript{$\dagger$} \citep{sridhar2025memerscalingmemoryrobot}; PaLM-E \citep{driess2023palm}; SpatialVLA\textsuperscript{$\dagger$} \citep{qu2025spatialvlaexploringspatialrepresentations}; Statler \citep{yoneda2024statler}; UP-VLA \citep{zhang2025up} \\
\textbf{R/T} & 3 & ChatVLA \citep{zhou2025chatvla}; LEO \citep{huang2024embodied}; NavGPT-2 \citep{zhou2024navgpt2} \\
\rowcolor{panoramarow}\textbf{R} & 2 & CogACT\textsuperscript{$\dagger$} \citep{li2024cogactfoundationalvisionlanguageactionmodel}; EnerVerse \citep{huang2026enerverse} \\
\textbf{R/T/C} & 2 & PhysObjects \citep{gao2024physically}; VoxPoser \citep{huang2023voxposer} \\
\rowcolor{panoramarow}\textbf{R/C} & 2 & Distilled Feature Fields \citep{shen2023distilled}; RoboPoint \citep{yuan2025robopoint} \\
\textbf{R/T/C/F} & 2 & Manipulate-Anything \citep{duan2025manipulate}; OneTwoVLA\textsuperscript{$\dagger$} \citep{lin2026onetwovlaunifiedvisionlanguageactionmodel} \\
\rowcolor{panoramarow}\textbf{R/C/F} & 1 & DexGraspVLA \citep{zhong2026dexgraspvla} \\
\end{coreinventorytable}

\begin{coreinventorytable}{Paper inventory for Action Orchestration.}{tab:core-inventory-action-orchestration}
\textbf{R/T/C/I} & 5 & Grounded Decoding \citep{huang2023grounded}; ProgPrompt \citep{singh2023progprompt}; SayCan\textsuperscript{$\dagger$} \citep{ahn2022icanisay}; Unified Agent\textsuperscript{$\dagger$} \citep{dipalo2023unifiedagentfoundationmodels}; Zero-Shot Planners \citep{huang2022language} \\
\rowcolor{panoramarow}\textbf{R/T/C/F/I} & 6 & DECKARD \citep{nottingham2023embodied}; DEPS \citep{wang2023describe}; LLM-DP\textsuperscript{$\dagger$} \citep{dagan2023dynamicplanningllm}; LLM-Planner \citep{song2023llm}; ReplanVLM \citep{mei2024replanvlm}; SEEA-R1 \citep{tian2026seea} \\
\textbf{R/T/C} & 1 & Octopus \citep{yang2024octopus} \\
\end{coreinventorytable}

\begin{coreinventorytable}{Paper inventory for Grounding Regulation.}{tab:core-inventory-grounding-regulation}
\textbf{R/T/C/I} & 6 & GRAPE\textsuperscript{$\dagger$} \citep{zhang2025grapegeneralizingrobotpolicy}; Hume\textsuperscript{$\dagger$} \citep{song2025humeintroducingsystem2thinking}; KnowNo \citep{ren2023robots}; Large Reward Models\textsuperscript{$\dagger$} \citep{wu2026largerewardmodelsgeneralizable}; RoboFuME \citep{yang2024robot}; VLM-Social-Nav \citep{song2024vlm} \\
\rowcolor{panoramarow}\textbf{R/T/C/F/I} & 11 & AHA\textsuperscript{$\dagger$} \citep{duan2024ahavisionlanguagemodeldetectingreasoning}; DROC \citep{zha2024distilling}; ELITE\textsuperscript{$\dagger$} \citep{wei2026eliteexperientiallearningintentaware}; EmbodiSkill\textsuperscript{$\dagger$} \citep{ju2026embodiskillskillawarereflectionselfevolving}; Evolvable Embodied Agent\textsuperscript{$\dagger$} \citep{wang2026evolvableembodiedagentrobotic}; Inner Monologue \citep{huang2023inner} \\
\textbf{R/T/C/F/I (cont.)} &  & Language to Rewards \citep{yu2023language}; Reflective Planning\textsuperscript{$\dagger$} \citep{hong2026learningtrialserrorsreflective}; SAIL\textsuperscript{$\dagger$} \citep{sato2026sailtesttimescalingincontext}; Text2Reward \citep{xie2024text2reward}; VLA-RL\textsuperscript{$\dagger$} \citep{lu2025vlarlmasterfulgeneralrobotic} \\
\rowcolor{panoramarow}\textbf{R/T/C/F} & 1 & APO \citep{xia2026human} \\
\end{coreinventorytable}

\begin{coreinventorytable}{Paper inventory for Execution Coupling.}{tab:core-inventory-execution-coupling}
\textbf{R/T/C/I} & 1 & Code as Policies \citep{liang2023code} \\
\rowcolor{panoramarow}\textbf{R/T/C/F/I} & 1 & ReKep \citep{huang2025rekep} \\
\textbf{R/T/I} & 1 & DexVLA \citep{wen2025dexvla} \\
\rowcolor{panoramarow}\textbf{R/T} & 1 & RT-2 \citep{zitkovich2023rt} \\
\end{coreinventorytable}

\FloatBarrier
\clearpage

\begin{strip}
\input{appendices/generated/d_evidence_profile_upset}
\end{strip}

\section{Joint Evidence Profiles in the Reviewed Corpus}
\label{app:evidence-profiles}

Marginal frequencies do not show which evidence operations occur together.
Figure~\ref{fig:evidence-profile-upset} reports the nine mutually exclusive R/T/C/F/I
profiles observed in the reviewed corpus.

The three most frequent profiles are RTCI (39 papers), RTCFI (26), and RTI (20).
Together they account for 85 of the 105 reviewed papers. All 30 F-positive papers also
satisfy C. This co-occurrence is descriptive under the stated coding rules and does not
imply that C is a logical prerequisite for F.

%% file: appendices/generated/d_evidence_profile_upset.tex
\centering
\begin{tikzpicture}[x=1cm,y=1cm]
  \tikzset{auditdot/.style={circle,minimum size=4.0mm,inner sep=0pt,line width=0.45pt}}
  \node[anchor=east,font=\bfseries\scriptsize] at (0.50,3.16) {Papers};
  \draw[black!32,line width=0.45pt] (0.72,0.95) -- (10.62,0.95);
  \fill[TakeawayLine!88] (0.84,0.95) rectangle (1.46,3.00);
  \node[font=\bfseries\scriptsize,anchor=south] at (1.15,3.04) {39};
  \fill[TakeawayLine!88] (1.97,0.95) rectangle (2.59,2.32);
  \node[font=\bfseries\scriptsize,anchor=south] at (2.28,2.36) {26};
  \fill[TakeawayLine!88] (3.10,0.95) rectangle (3.72,2.00);
  \node[font=\bfseries\scriptsize,anchor=south] at (3.41,2.04) {20};
  \fill[TakeawayLine!88] (4.23,0.95) rectangle (4.85,1.21);
  \node[font=\bfseries\scriptsize,anchor=south] at (4.54,1.25) {5};
  \fill[TakeawayLine!88] (5.36,0.95) rectangle (5.98,1.16);
  \node[font=\bfseries\scriptsize,anchor=south] at (5.67,1.20) {4};
  \fill[TakeawayLine!88] (6.49,0.95) rectangle (7.11,1.16);
  \node[font=\bfseries\scriptsize,anchor=south] at (6.80,1.20) {4};
  \fill[TakeawayLine!88] (7.62,0.95) rectangle (8.24,1.11);
  \node[font=\bfseries\scriptsize,anchor=south] at (7.93,1.15) {3};
  \fill[TakeawayLine!88] (8.75,0.95) rectangle (9.37,1.11);
  \node[font=\bfseries\scriptsize,anchor=south] at (9.06,1.15) {3};
  \fill[TakeawayLine!88] (9.88,0.95) rectangle (10.50,1.00);
  \node[font=\bfseries\scriptsize,anchor=south] at (10.19,1.04) {1};
  \node[anchor=east,font=\bfseries\scriptsize] at (0.50,0.28) {R};
  \node[anchor=east,font=\bfseries\scriptsize] at (0.50,-0.28) {T};
  \node[anchor=east,font=\bfseries\scriptsize] at (0.50,-0.84) {C};
  \node[anchor=east,font=\bfseries\scriptsize] at (0.50,-1.4) {F};
  \node[anchor=east,font=\bfseries\scriptsize] at (0.50,-1.96) {I};
  \node[anchor=south,font=\bfseries\scriptsize] at (-1.18,0.59) {Papers};
  \draw[black!32,line width=0.45pt] (-0.1,0.50) -- (-0.1,-2.18);
  \fill[TakeawayLine!72] (-2.25,0.14) rectangle (-0.1,0.42);
  \node[anchor=east,font=\bfseries\scriptsize] at (-2.33,0.28) {105};
  \fill[TakeawayLine!72] (-2.09,-0.42) rectangle (-0.1,-0.14);
  \node[anchor=east,font=\bfseries\scriptsize] at (-2.17,-0.28) {97};
  \fill[TakeawayLine!72] (-1.66,-0.98) rectangle (-0.1,-0.70);
  \node[anchor=east,font=\bfseries\scriptsize] at (-1.74,-0.84) {76};
  \fill[TakeawayLine!72] (-0.71,-1.54) rectangle (-0.1,-1.26);
  \node[anchor=east,font=\bfseries\scriptsize] at (-0.79,-1.4) {30};
  \fill[TakeawayLine!72] (-1.84,-2.10) rectangle (-0.1,-1.82);
  \node[anchor=east,font=\bfseries\scriptsize] at (-1.92,-1.96) {85};
  \draw[TakeawayLine!72,line width=1.05pt] (1.15,-1.96) -- (1.15,0.28);
  \node[auditdot,draw=TakeawayLine,fill=TakeawayLine] at (1.15,0.28) {};
  \node[auditdot,draw=TakeawayLine,fill=TakeawayLine] at (1.15,-0.28) {};
  \node[auditdot,draw=TakeawayLine,fill=TakeawayLine] at (1.15,-0.84) {};
  \node[auditdot,draw=black!24,fill=white] at (1.15,-1.4) {};
  \node[auditdot,draw=TakeawayLine,fill=TakeawayLine] at (1.15,-1.96) {};
  \node[font=\scriptsize,anchor=north] at (1.15,-2.32) {RTCI};
  \draw[TakeawayLine!72,line width=1.05pt] (2.28,-1.96) -- (2.28,0.28);
  \node[auditdot,draw=TakeawayLine,fill=TakeawayLine] at (2.28,0.28) {};
  \node[auditdot,draw=TakeawayLine,fill=TakeawayLine] at (2.28,-0.28) {};
  \node[auditdot,draw=TakeawayLine,fill=TakeawayLine] at (2.28,-0.84) {};
  \node[auditdot,draw=TakeawayLine,fill=TakeawayLine] at (2.28,-1.4) {};
  \node[auditdot,draw=TakeawayLine,fill=TakeawayLine] at (2.28,-1.96) {};
  \node[font=\scriptsize,anchor=north] at (2.28,-2.32) {RTCFI};
  \draw[TakeawayLine!72,line width=1.05pt] (3.41,-1.96) -- (3.41,0.28);
  \node[auditdot,draw=TakeawayLine,fill=TakeawayLine] at (3.41,0.28) {};
  \node[auditdot,draw=TakeawayLine,fill=TakeawayLine] at (3.41,-0.28) {};
  \node[auditdot,draw=black!24,fill=white] at (3.41,-0.84) {};
  \node[auditdot,draw=black!24,fill=white] at (3.41,-1.4) {};
  \node[auditdot,draw=TakeawayLine,fill=TakeawayLine] at (3.41,-1.96) {};
  \node[font=\scriptsize,anchor=north] at (3.41,-2.32) {RTI};
  \draw[TakeawayLine!72,line width=1.05pt] (4.54,-0.28) -- (4.54,0.28);
  \node[auditdot,draw=TakeawayLine,fill=TakeawayLine] at (4.54,0.28) {};
  \node[auditdot,draw=TakeawayLine,fill=TakeawayLine] at (4.54,-0.28) {};
  \node[auditdot,draw=black!24,fill=white] at (4.54,-0.84) {};
  \node[auditdot,draw=black!24,fill=white] at (4.54,-1.4) {};
  \node[auditdot,draw=black!24,fill=white] at (4.54,-1.96) {};
  \node[font=\scriptsize,anchor=north] at (4.54,-2.32) {RT};
  \node[auditdot,draw=TakeawayLine,fill=TakeawayLine] at (5.67,0.28) {};
  \node[auditdot,draw=black!24,fill=white] at (5.67,-0.28) {};
  \node[auditdot,draw=black!24,fill=white] at (5.67,-0.84) {};
  \node[auditdot,draw=black!24,fill=white] at (5.67,-1.4) {};
  \node[auditdot,draw=black!24,fill=white] at (5.67,-1.96) {};
  \node[font=\scriptsize,anchor=north] at (5.67,-2.32) {R};
  \draw[TakeawayLine!72,line width=1.05pt] (6.80,-0.84) -- (6.80,0.28);
  \node[auditdot,draw=TakeawayLine,fill=TakeawayLine] at (6.80,0.28) {};
  \node[auditdot,draw=TakeawayLine,fill=TakeawayLine] at (6.80,-0.28) {};
  \node[auditdot,draw=TakeawayLine,fill=TakeawayLine] at (6.80,-0.84) {};
  \node[auditdot,draw=black!24,fill=white] at (6.80,-1.4) {};
  \node[auditdot,draw=black!24,fill=white] at (6.80,-1.96) {};
  \node[font=\scriptsize,anchor=north] at (6.80,-2.32) {RTC};
  \draw[TakeawayLine!72,line width=1.05pt] (7.93,-0.84) -- (7.93,0.28);
  \node[auditdot,draw=TakeawayLine,fill=TakeawayLine] at (7.93,0.28) {};
  \node[auditdot,draw=black!24,fill=white] at (7.93,-0.28) {};
  \node[auditdot,draw=TakeawayLine,fill=TakeawayLine] at (7.93,-0.84) {};
  \node[auditdot,draw=black!24,fill=white] at (7.93,-1.4) {};
  \node[auditdot,draw=black!24,fill=white] at (7.93,-1.96) {};
  \node[font=\scriptsize,anchor=north] at (7.93,-2.32) {RC};
  \draw[TakeawayLine!72,line width=1.05pt] (9.06,-1.40) -- (9.06,0.28);
  \node[auditdot,draw=TakeawayLine,fill=TakeawayLine] at (9.06,0.28) {};
  \node[auditdot,draw=TakeawayLine,fill=TakeawayLine] at (9.06,-0.28) {};
  \node[auditdot,draw=TakeawayLine,fill=TakeawayLine] at (9.06,-0.84) {};
  \node[auditdot,draw=TakeawayLine,fill=TakeawayLine] at (9.06,-1.4) {};
  \node[auditdot,draw=black!24,fill=white] at (9.06,-1.96) {};
  \node[font=\scriptsize,anchor=north] at (9.06,-2.32) {RTCF};
  \draw[TakeawayLine!72,line width=1.05pt] (10.19,-1.40) -- (10.19,0.28);
  \node[auditdot,draw=TakeawayLine,fill=TakeawayLine] at (10.19,0.28) {};
  \node[auditdot,draw=black!24,fill=white] at (10.19,-0.28) {};
  \node[auditdot,draw=TakeawayLine,fill=TakeawayLine] at (10.19,-0.84) {};
  \node[auditdot,draw=TakeawayLine,fill=TakeawayLine] at (10.19,-1.4) {};
  \node[auditdot,draw=black!24,fill=white] at (10.19,-1.96) {};
  \node[font=\scriptsize,anchor=north] at (10.19,-2.32) {RCF};
  \node[auditdot,draw=TakeawayLine,fill=TakeawayLine] at (2.25,-3.03) {};
  \node[anchor=west,font=\scriptsize] at (2.53,-3.03) {operation observed};
  \node[auditdot,draw=black!24,fill=white] at (5.35,-3.03) {};
  \node[anchor=west,font=\scriptsize] at (5.63,-3.03) {not observed};
\end{tikzpicture}

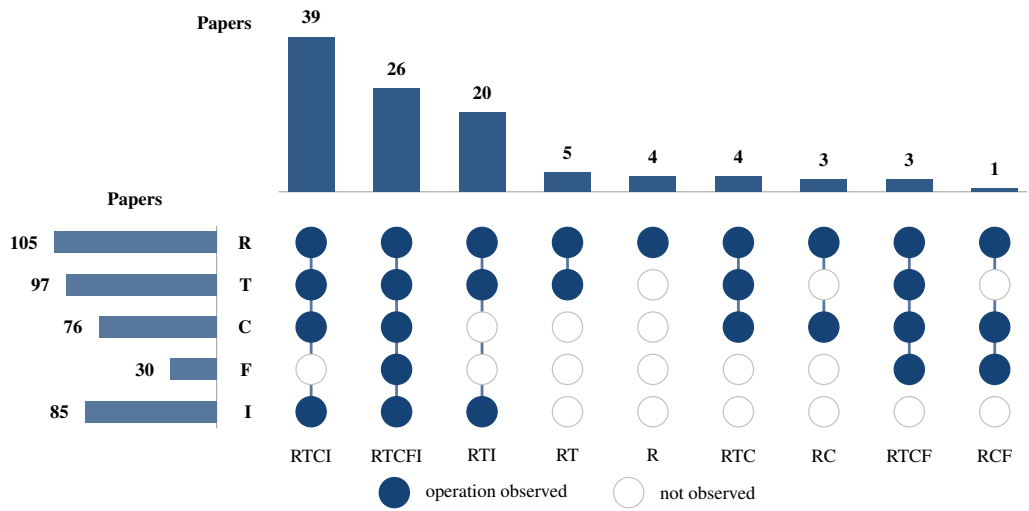
\captionof{figure}{Exact joint evidence profiles in the reviewed corpus. Bars above the columns
give the number of papers in each mutually exclusive profile, and horizontal bars at
left give the marginal count for each operation. Filled dots mark observed operations.
Connecting segments group operations within a profile and do not indicate order or
dependence. R appears in every profile because route traceability is an inclusion
requirement.}
\label{fig:evidence-profile-upset}